# A Deep Learning Approach for Similar Languages, Varieties and Dialects

**Vidya Prasad K, Akarsh S, Vinayakumar R, Soman KP**
**Center for Computational Engineering and Networking(CEN), Amrita School of Engineering, Coimbatore**
**Email:** vidya.prasad.k565@gmail.com, akarshsoman@gmail.com

**Abstract**

Deep learning mechanisms are prevailing approaches in recent days for the various tasks in natural language processing, speech recognition, image processing and many others. To leverage this we use deep learning based mechanism specifically Bidirectional- Long Short-Term Memory (B-LSTM) for the task of dialectic identification in Arabic and German broadcast speech and Long Short-Term Memory (LSTM) for discriminating between similar Languages. Two unique B-LSTM models are created using the Large-vocabulary Continuous Speech Recognition (LVCSR) based lexical features and a fixed length of 400 per utterance bottleneck features generated by i-vector framework. These models were evaluated on the VarDial 2017 datasets for the tasks Arabic, German dialect identification with dialects of Egyptian, Gulf, Levantine, North African, and MSA for Arabic and Basel, Bern, Lucerne, and Zurich for German. Also for the task of Discriminating between Similar Languages like Bosnian, Croatian and Serbian. The B-LSTM model showed accuracy of 0.246 on lexical features and accuracy of 0.577 bottleneck features of i-Vector framework. The performance obtained by B-LSTM models is considerable less, mainly due to the reason that the .

**Keywords:** Dialect Identification, Deep Learning; Recurrent Neural Network (RNN), Bidirectional Recurrent Neural Network (B-RNN), Bidirectional Long-Short Term Memory (B-LSTM)

## 1 Introduction

For the past three decades, Language Identification (LI) has remained a vivid area of research like focusing on identifying the correct language for given speech signal or text file. Dialect Identification (DID) is a subnet of LI that has been widely studying by speech research communities in recent days [1]. DI is the task of identifying the dialects of a speech signal considering a particular language. This is considered as more challenging task in comparison to the LI due to the fact that the dialects resemble each other in the same language. A better dialect identification system lessens the word error rates of Automatic Speech Recognition (ASR) by recognizing the dialectal chunks of un-transcribed mixed-speech.

Egyptian (EGY), North African or Maghrebi (NOR), Gulf or Arabian Peninsula (GLF), Levantine (LAV), and Modern Standard Arabic (MSA) are five regional language categories of Arabic dialects. Commercial products for Automatic Speech Recognition (ASR) existing in markets are merely achieving low error rates in recognizing MSA. While, in the case of recognizing vernacular phrases, the commercial systems show higher error rates or in some cases solely the system fails [8]. The dialects in Arabic languages are unique and the task of identifying the dialects of various Arabic languages is similar to the language identification task.German dialects being less studied than Arabic, here four Swiss-German dialects: Basel(BS), Bern(BE), Lucerne(LU), and Zurich(ZH) were also identified using the same DI architecture.

Recognizing the dialects of regional languages of a speaker allows ASR to adopt the various contextual and linguistic features and accordingly enhances the performance of speech recognition systems. Additionally, DI provides the origin and ethnicity features of native speaker that becomes useful in identifying the regional speaker identification.

Language Identification also consists of discriminating between similar languages. In this task, the textual data from newspapers for different languages which are similar in nature like Malay(my) and Indonesian(id); Bosnian(bs), Croatian(hr) and Serbian(sr); Persian(fa-IR) and Dari(fa-AF); Canadian(fr-CA) and Hexagonal French(fr-FR); Brazilian(pt-BR) and European Portuguese(pt-PT); Argentinian(es-AR), Peninsular(es-ES), and Peruvian Spanish(es-PE).

The model proposed here was applied for these 3 different tasks of VarDial 2017: Discriminating Similar Languages(DSL), Arabic Dialect Identification(ADI) and German Dialect Identification(GDI).

The rest of the paper is organized as follows; section 2 discusses the related work on dialect identification and deep learning mechanisms. Section 3 provides the necessary mathematical background and the concepts of LSTM and bidirectional long short term memory (B-LSTM). Section 4 displays the evaluation of the LSTM architecture for discrimination of similar languages and its results. Sections 5 and 6 contain the evaluation of B-LSTM model for Arabic and German Dialect Identification. Finally the conclusion is placed at the last section of the a paper.

## 2 Related Work

Distinguishing two similar languages is a hurdle for Language Identification systems especially when the languages are nearly related.This area has been explored recently like the works by on South-Slavic languages[24], Mandarin varieties in Singapore,Taiwan and China [25], Malay and Indonesian languages [26]. Inspired by such works, a lot of shared task competitions give importance to such tasks.

Arabic Dialect Identification for NLP is a new branch and significant amount of work has been done in this area for both speech[9-11] as well as textual forms[13]. A.Shoufan and S.Al-Ameri discusses in detail the different approaches done for Arabic Dialect identification[12] till 2015. It can be used as a source to get information about the relevant works on Arabic dialect identification using both speech and text data. An approach by taking p-grams on the transcribed speech data was done by [20] and [21]. In our work, the DID is done for speech transcripts and acoustic features provided in the VarDial 2017 subtask. The acoustic features provided are i-Vectors obtained using Bottle Neck Features. Bahari.et.al uses Identical Vectors or i-Vectors to extract features like accents and dialects in speech processing for effective results[14]. Kernel Ridge Regression (KRR), Support Vector Machine (SVM) and ensemble methods give the best results [23].

Similarly German Dialect Identification was another subtask for the competition. Works on German Dialect Identification are lesser in number when compared to Arabic. One of the initial works use n-gram method at a character level for this[18].Another very different approach included recognising the high frequency words in the data[19].The dataset that we use here consists of speech transcripts provided by the VarDial 2017 subtask taken from ArchiMob Corpus of Spoken Swiss German released in 2016 [22]. The best results in this task was shown by a meta-classifier based on SVM classifiers [23].

Deep learning mechanisms have established as a significant methods in recent days with solving the long standing artificial intelligence tasks in natural language processing, image processing, speech recognition and many others [1]. In recent days, application of deep learning architectures are leveraged for cyber security [34], [35], [36], [37], [38] and genomics and proteomics [39]. R.Zazo et.al used LSTM in their work for Language Identification and this model is compared with an i-Vector based system. The LSTM model shows better performance than the latter [27]. Long short-term memory (LSTM) is one of significant methods of deep learning that focusing on learning long-range temporal dependencies in large sequences of arbitrary length [2, 3, 4]. LSTM have established as a promising approach for sequence data modeling by solving the tasks such as machine translation [5] and many others [6]. Hence in this work, we use a LSTM model the subtask of Discriminating Similar Languages (DSL) while we use Bidirectional LSTM for Arabic Dialect Identification(ADI) and German Dialect Identification(GDI) .

## 3 Background

This section discusses the network architectures:LSTM and B-LSTM implemented by us and particularly B-LSTM concisely.

### 3.1 Network Design

Recurrent neural network is mostly used in sequence modelling problems in the domain of natural language processing (NLP) [29, 30, 31, 32, 33]. The significant issue of RNN is that vanishing and exploding gradient [28]. Long short-term memory (LSTM) is one of significant methods of deep learning that focusing on learning long-range temporal dependencies in large sequences of arbitrary length [2, 3, 4]. LSTM have established as a promising approach for sequence data modeling by solving the tasks such as machine translation [5] and many others [6].

$$i_t = \sigma(w_{xi}x_t + w_{hi}h_{t-1} + w_{ci}c_{t-1} + b_i)$$

$$f_t = \sigma(w_{xf}x_t + w_{hf}h_{t-1} + w_{cf}c_{t-1} + b_f)$$

$$c_t = f_t \odot c_{t-1} + i_t \odot \tanh(w_{xc}x_t + w_{hc}h_{t-1} + b_c)$$

$$o_t = SG(w_{xo}x_t + w_{ho}h_{t-1} + w_{co}c_t + b_o)$$

$$h_t = o_t \odot \tanh(c_t)$$

Where $i$ , $o$ , $f$ , $c$ term denotes the input gate output gate, forget gate and a memory cell respectively. Conventional RNN only facilitate to learn the past dependencies. Capturing future context in sequences is important in speech recognition due to the fact that the transcription of the whole utterance is done only once and for the better performance of the system, future context can be exploited as well. Bidirectional recurrent neural network (B-RNN) achieves this by capturing the long-range dependencies of temporal patterns in sequence data in both the forward and backward directions. As shown in Fig 2, given an input sequence $x = (x_1, x_{2,} ...., x_T)$ BRNN maps input sequence to forward hidden states and backward hidden states and forward hidden states and backward hidden states to output sequences from $t = 1$ to $T$ by iterating the following recursive equations.

$$hf_t = SG(w_{xh}x_t + w_{hh}h_{t-1} + b_h) \text{ (Forward direction)}$$

$$hb_t = SG(w_{xh}x_t + w_{hh}h_{t-1} + b_h) \text{ (Backward direction)}$$

$$o_t = SG(w_{hfo}hf_t + w_{hbo}hb_t + b_o)$$

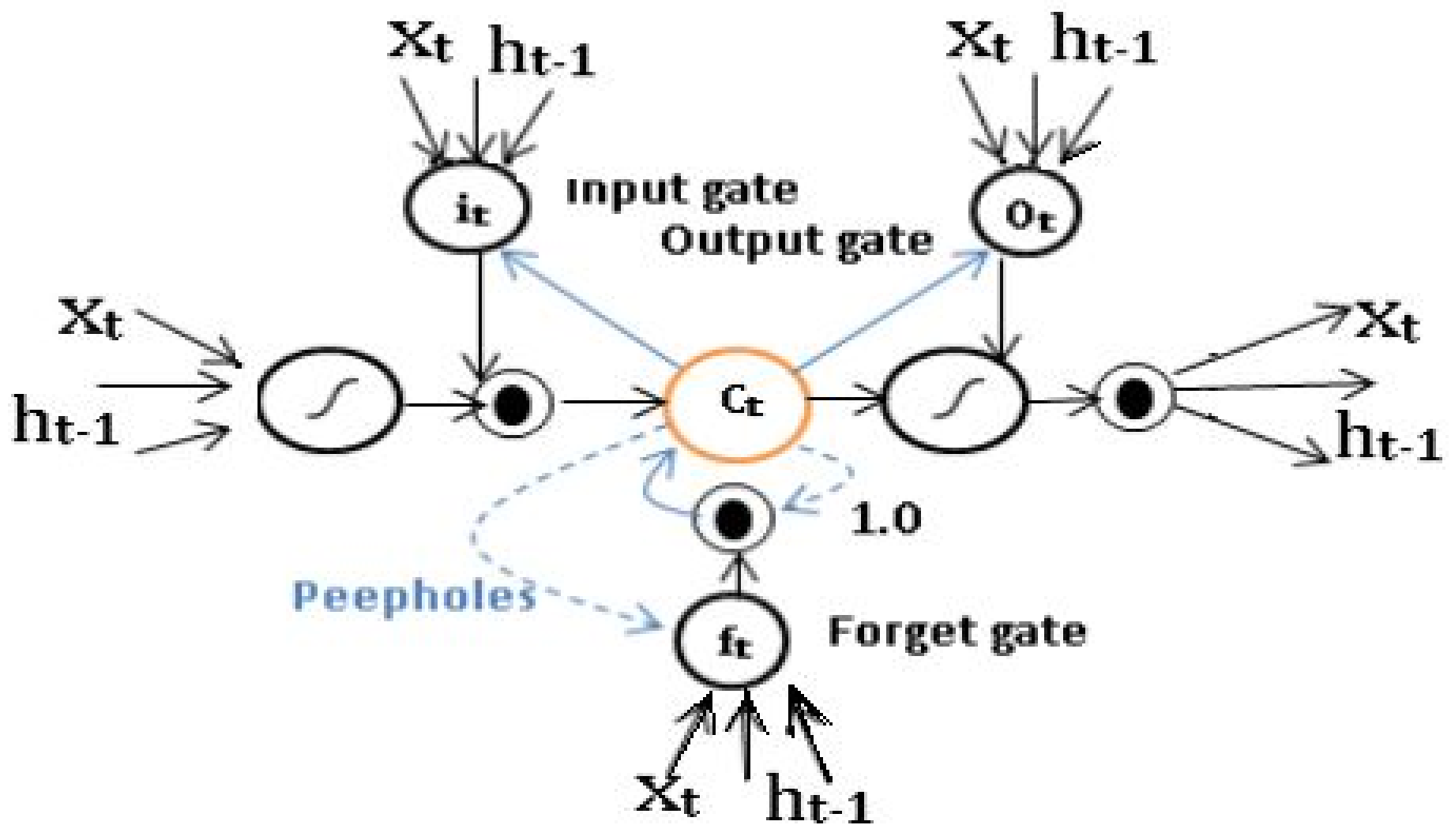

**Fig 1** LSTM memory cell

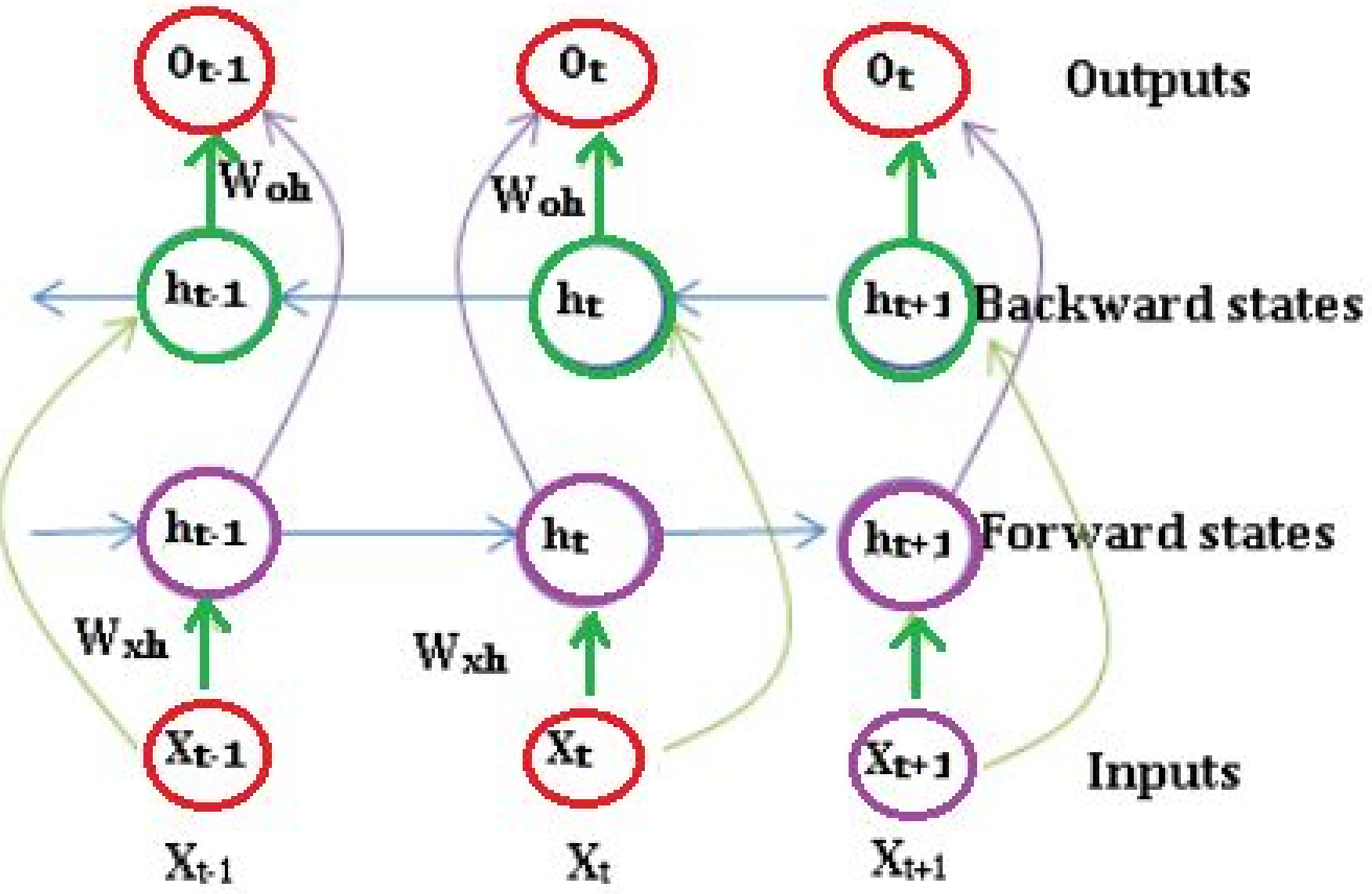

**Fig 2** Bidirectional Recurrent Neural Network

B-RNN is composed of conventional RNN and LSTM that facilitate to learn the dependencies in both forward and backward directions [7].

## 4. Experiments

This section includes experimental analysis on 3 different tasks, Discriminating between Similar Languages, Arabic Dialect Identification and German Dialect Identification. These tasks are part of VarDial 2017 shared task [23].

### 4.1 Discriminating Similar Languages

#### 4.1.1 Description of data set

The dataset consists of 14 languages in textual form, the similar languages being:

- Bosnian,Croatian,Serbian
- Indonesian,Malay
- Brazilian Portuguese,European Portuguese
- Argentine Spanish,Castilian Spanish,Peruvian Spanish
- Canadian French,Hexagonal French
- Persian,Dari

Each dataset has 22,000 excerpts of which 18,000 make the training data, 2,000 make the development set and 2,000 for testing.

#### 4.1.2 Methodology

Here we use the deep learning architecture LSTM for distinguishing the similar languages. The first run results are that of character based models. The second run results are from word level based models.

#### 4.1.3 Results

| Team | Test Set | Track | Run | Accuracy | F1 (micro) | F1 (macro) | F1 (weighted) |
|---|---|---|---|---|---|---|---|
| deepCybErNet | DSL | closed | Character based | 0.205 | 0.205 | 0.202 | 0.202 |
| deepCybErNet | DSL | closed | Word Based | 0.195 | 0.195 | 0.186 | 0.186 |

**Table 1.** Statistical Measures for Lexical and i-Vector based Features

| | **hr** | **bs** | **sr** | **es-ar** | **es-es** | **es-pe** | **fa-af** | **fa-ir** | **fr-ca** | **fr-fr** | **id** | **my** | **pt-br** | **pt-pt** |
|---|---|---|---|---|---|---|---|---|---|---|---|---|---|---|
| **hr** | 381 | 142 | 296 | 2 | 3 | 91 | 14 | 1 | 7 | 22 | 6 | 11 | 20 | 4 |
| **bs** | 328 | 154 | 271 | 2 | 7 | 136 | 21 | 1 | 9 | 19 | 6 | 11 | 31 | 4 |
| **sr** | 406 | 121 | 238 | 11 | 14 | 129 | 9 | 1 | 5 | 23 | 2 | 13 | 24 | 4 |
| **es-ar** | 2 | 19 | 25 | 170 | 225 | 114 | 5 | 5 | 160 | 131 | 19 | 11 | 19 | 95 |
| **es-es** | 3 | 15 | 25 | 184 | 271 | 82 | 2 | 5 | 171 | 106 | 22 | 7 | 11 | 96 |
| **es-pe** | 10 | 32 | 25 | 120 | 133 | 199 | 4 | 3 | 233 | 149 | 12 | 0 | 7 | 73 |
| **fa-af** | 6 | 11 | 5 | 18 | 47 | 32 | 250 | 25 | 138 | 114 | 214 | 125 | 1 | 14 |
| **fa-ir** | 4 | 10 | 9 | 27 | 45 | 33 | 201 | 78 | 109 | 84 | 262 | 122 | 3 | 13 |
| **fr-ca** | 1 | 1 | 78 | 70 | 15 | 155 | 86 | 29 | 207 | 94 | 14 | 24 | 25 | 201 |
| **fr-fr** | 3 | 1 | 89 | 60 | 26 | 156 | 92 | 37 | 154 | 139 | 20 | 32 | 38 | 153 |
| **id** | 5 | 2 | 12 | 36 | 71 | 82 | 104 | 28 | 87 | 97 | 198 | 158 | 38 | 82 |
| **my** | 3 | 2 | 6 | 31 | 47 | 87 | 176 | 17 | 88 | 100 | 170 | 165 | 28 | 80 |
| **pt-br** | 67 | 6 | 18 | 88 | 23 | 101 | 55 | 100 | 104 | 62 | 17 | 5 | 115 | 239 |
| **pt-pt** | 47 | 7 | 9 | 83 | 29 | 117 | 49 | 86 | 85 | 53 | 15 | 6 | 115 | 299 |

**Table 2.** Confusion Matrix for Character Based Model

| | **hr** | **bs** | **sr** | **es-ar** | **es-es** | **es-pe** | **fa-af** | **fa-ir** | **fr-ca** | **fr-fr** | **id** | **my** | **pt-br** | **pt-pt** |
|---|---|---|---|---|---|---|---|---|---|---|---|---|---|---|
| **hr** | 486 | 133 | 174 | 1 | 2 | 105 | 10 | 5 | 0 | 21 | 11 | 9 | 39 | 4 |
| **bs** | 470 | 134 | 139 | 4 | 5 | 120 | 9 | 0 | 0 | 32 | 10 | 24 | 47 | 6 |
| **sr** | 513 | 127 | 91 | 11 | 2 | 145 | 7 | 2 | 0 | 26 | 10 | 19 | 44 | 3 |
| **es-ar** | 2 | 22 | 7 | 221 | 230 | 84 | 9 | 16 | 37 | 180 | 36 | 29 | 40 | 87 |
| **es-es** | 8 | 15 | 6 | 241 | 222 | 101 | 6 | 9 | 46 | 158 | 35 | 27 | 43 | 83 |
| **es-pe** | 13 | 43 | 5 | 158 | 117 | 198 | 13 | 8 | 61 | 234 | 19 | 13 | 26 | 92 |
| **fa-af** | 7 | 6 | 3 | 43 | 91 | 20 | 80 | 71 | 22 | 120 | 271 | 223 | 22 | 21 |
| **fa-ir** | 3 | 10 | 4 | 44 | 53 | 15 | 71 | 131 | 30 | 85 | 331 | 185 | 29 | 9 |
| **fr-ca** | 8 | 2 | 33 | 68 | 34 | 190 | 59 | 87 | 56 | 132 | 45 | 50 | 131 | 105 |
| **fr-fr** | 7 | 6 | 39 | 56 | 37 | 133 | 68 | 104 | 47 | 177 | 59 | 64 | 117 | 86 |
| **id** | 6 | 8 | 0 | 30 | 37 | 70 | 105 | 32 | 6 | 121 | 279 | 198 | 37 | 71 |
| **my** | 7 | 3 | 3 | 28 | 24 | 79 | 168 | 16 | 11 | 138 | 250 | 168 | 38 | 67 |
| **pt-br** | 43 | 12 | 3 | 88 | 15 | 64 | 67 | 132 | 26 | 55 | 49 | 31 | 158 | 257 |
| **pt-pt** | 25 | 10 | 5 | 85 | 20 | 70 | 37 | 113 | 15 | 49 | 41 | 42 | 158 | 330 |

**Table 3.** Confusion Matrix for Word- Based model

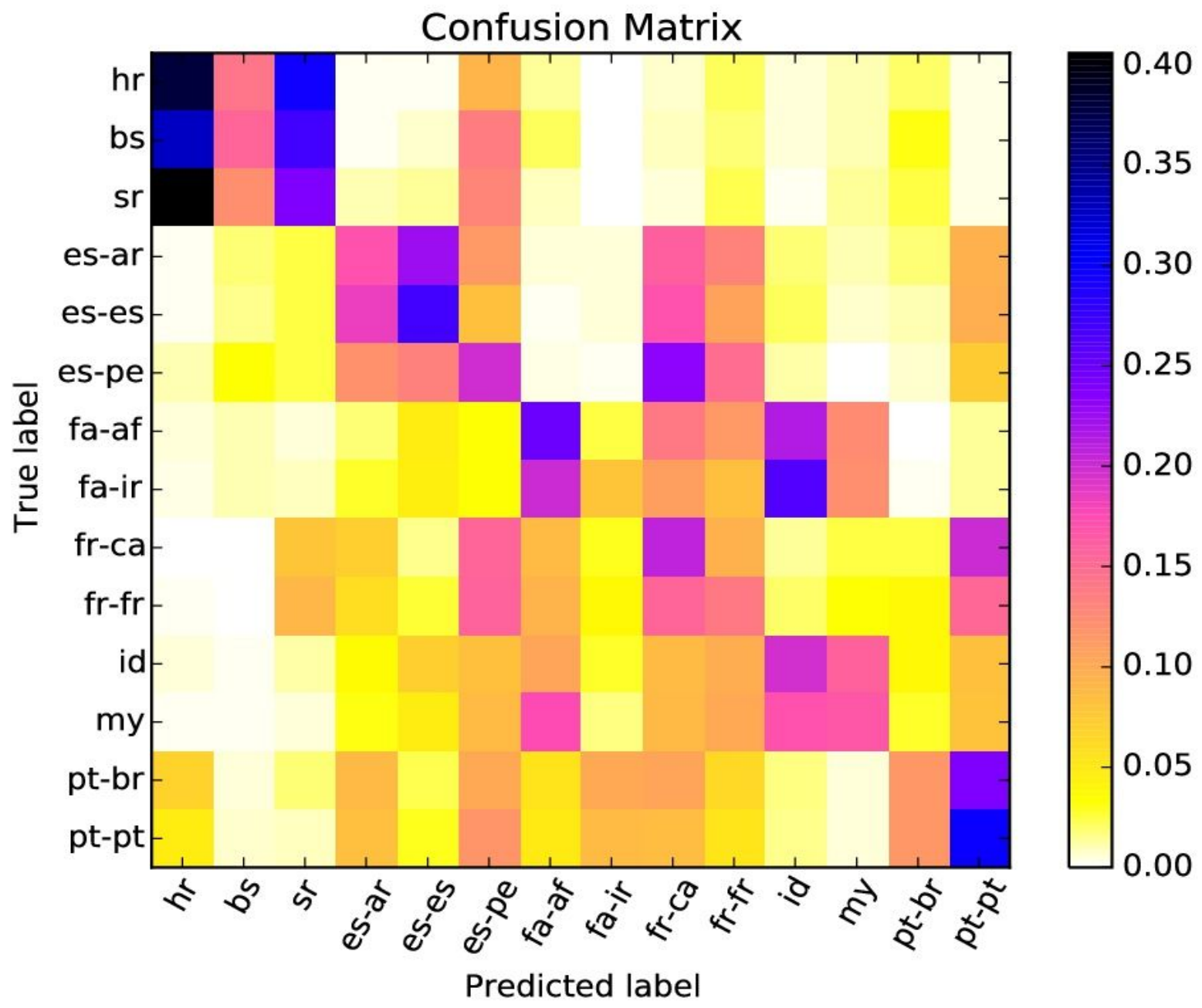

**Fig 3.** Confusion matrix for Character-based model

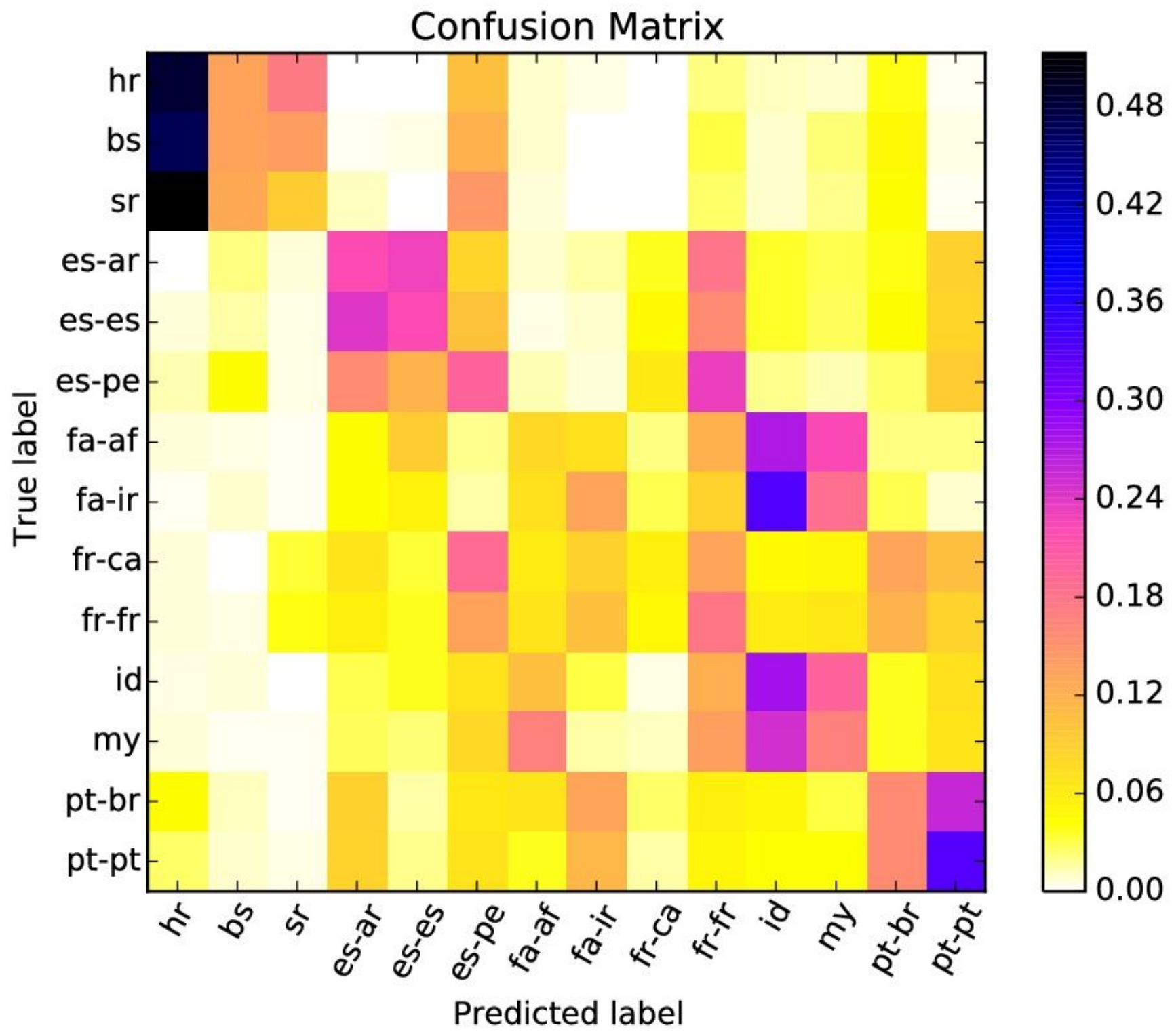

**Fig 4.** Confusion matrix for Word-based model

## 5. Experiments

### 5.1 Arabic Dialect Identification

#### 5.1.1 Description of data set

The dataset consists of both acoustic features and speech transcripts.

- The acoustic features are i-Vectors obtained by training on 60 hours of speech data using Bottle Neck Features
- The speech transcripts were generated by training 1200+ speech hours and 110+million words for acoustic and language modelling respectively.

#### 5.1.2 Methodology

For Arabic Dialect Identification, we adopted the variant of LSTM architecture; Bidirectional LSTM [3]. The first run result is based on the lexical features generated using LVCSR. The second run result is based on i-vector based on bottleneck features.

#### 5.1.3 Results

| Team | Test Set | Track | Run | Accuracy | F1 (micro) | F1 (macro) | F1 (weighted) |
|---|---|---|---|---|---|---|---|
| **deepCybErNet** | ADI | closed | lexical features | 0.246 | 0.246 | 0.204 | 0.208 |
| **deepCybErNet** | ADI | closed | i-vector features | 0.577 | 0.577 | 0.577 | 0.574 |

**Table 4.**Statistical Measures for Lexical and i-Vector based Features

| | egy | glf | lav | msa | nor |
|---|---|---|---|---|---|
| **egy** | 117 | 0 | 35 | 106 | 44 |
| **glf** | 108 | 0 | 19 | 97 | 26 |
| **lav** | 138 | 0 | 36 | 101 | 59 |
| **msa** | 109 | 0 | 6 | 145 | 2 |
| **nor** | 131 | 1 | 32 | 111 | 69 |

**Table 5.** Confusion Matrix for Lexical Features

| | egy | glf | lav | msa | nor |
|---|---|---|---|---|---|
| **egy** | 201 | 20 | 49 | 22 | 10 |
| **glf** | 28 | 107 | 77 | 30 | 8 |
| **lav** | 71 | 33 | 183 | 26 | 21 |
| **msa** | 20 | 10 | 16 | 211 | 5 |
| **nor** | 62 | 19 | 77 | 27 | 159 |

**Table 6.** Confusion Matrix for i-Vector Features

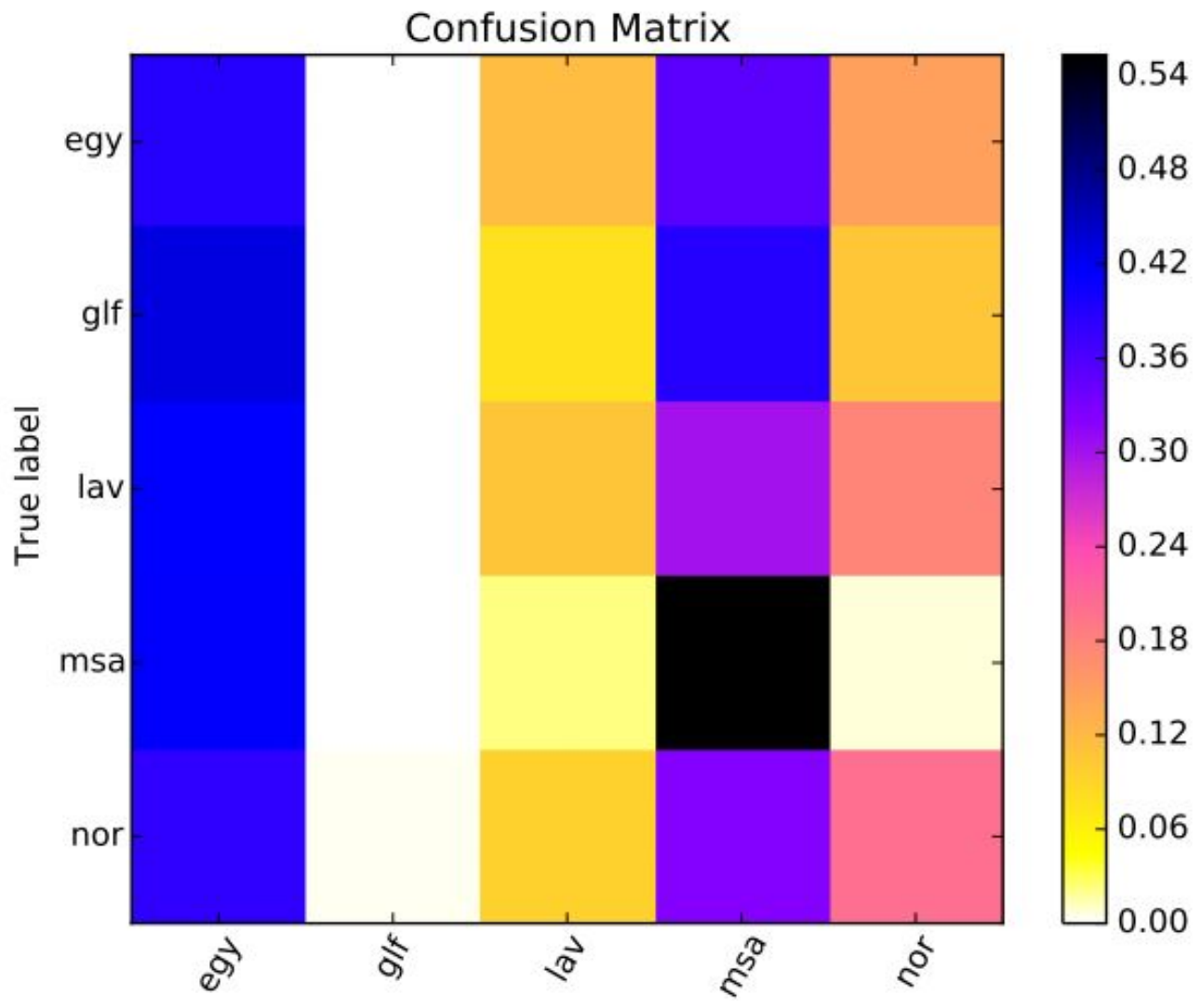

**Fig 5.** Confusion matrix for lexical features

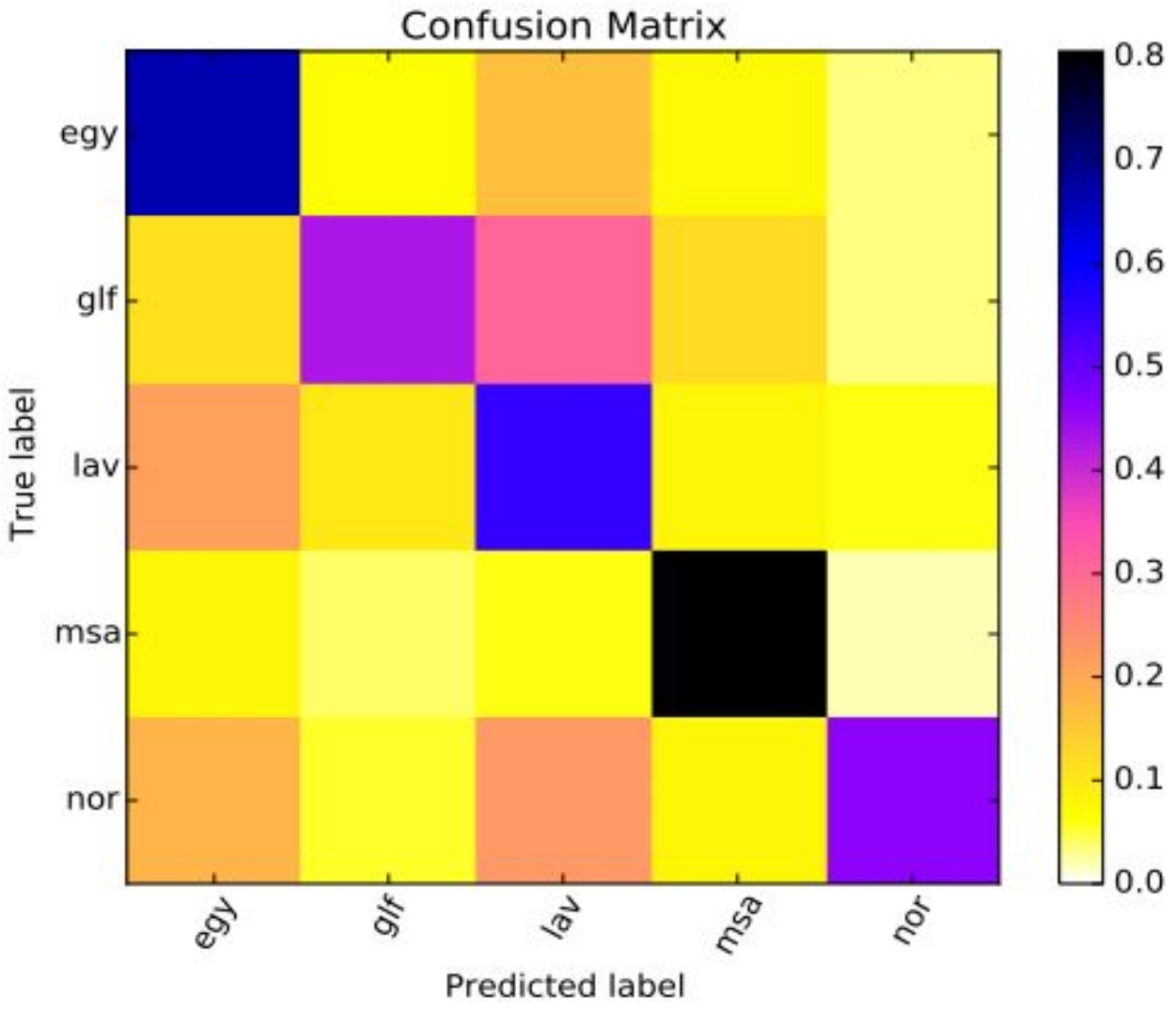

**Fig 6.** Confusion matrix for i-vector features

## 6 Experiments

### 5.1 German Dialect Identification

#### 6.1.1. Description of data set

The training and test sets were taken from the Spoken Swiss German from the ArchiMob corpus [22]. It has the transcriptions of 34 different oral interviews in different dialects of Swiss German.

### 6.1.2 Methodology

In this work, we applied the Bidirectional LSTM architecture [3] for German Dialect Identification. The first run a result is based on the character based models. The second run a result is based on word level based models.

### 6.1.3 Results

| Team | Test Set | Track | Run | Accuracy | F1 (micro) | F1 (macro) | F1 (weighted) |
|---|---|---|---|---|---|---|---|
| **deepCybErNet** | GDI | closed | lexical features | 0.263 | 0.263 | 0.264 | 0.263 |
| **deepCybErNet** | GDI | closed | i-vector features | 0.255 | 0.255 | 0.256 | 0.256 |

**Table 7**.Statistical Measures for Lexical and i-Vector based Features

| | **be** | **bs** | **lu** | **zh** |
|---|---|---|---|---|
| **be** | 259 | 237 | 251 | 159 |
| **bs** | 193 | 221 | 257 | 268 |
| **lu** | 248 | 287 | 204 | 177 |
| **zh** | 192 | 241 | 171 | 273 |

**Table 8.** Confusion Matrix for Lexical Features

| | **be** | **bs** | **lu** | **zh** |
|---|---|---|---|---|
| **be** | 251 | 257 | 251 | 147 |
| **bs** | 222 | 242 | 270 | 205 |
| **lu** | 256 | 288 | 220 | 152 |
| **zh** | 238 | 255 | 168 | 216 |

**Table 9.** Confusion Matrix for i-Vector Features

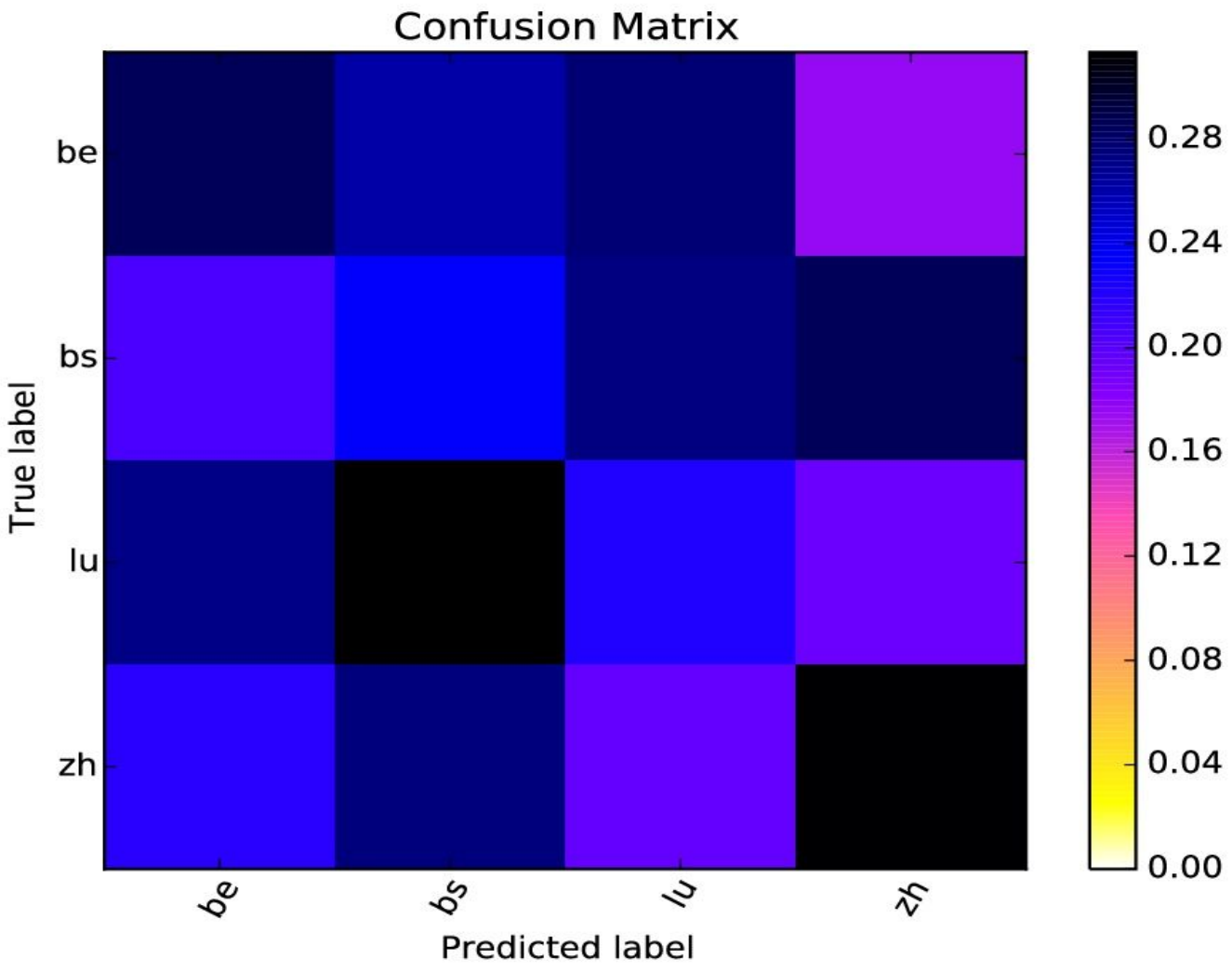

**Fig 7.** Confusion matrix for lexical features

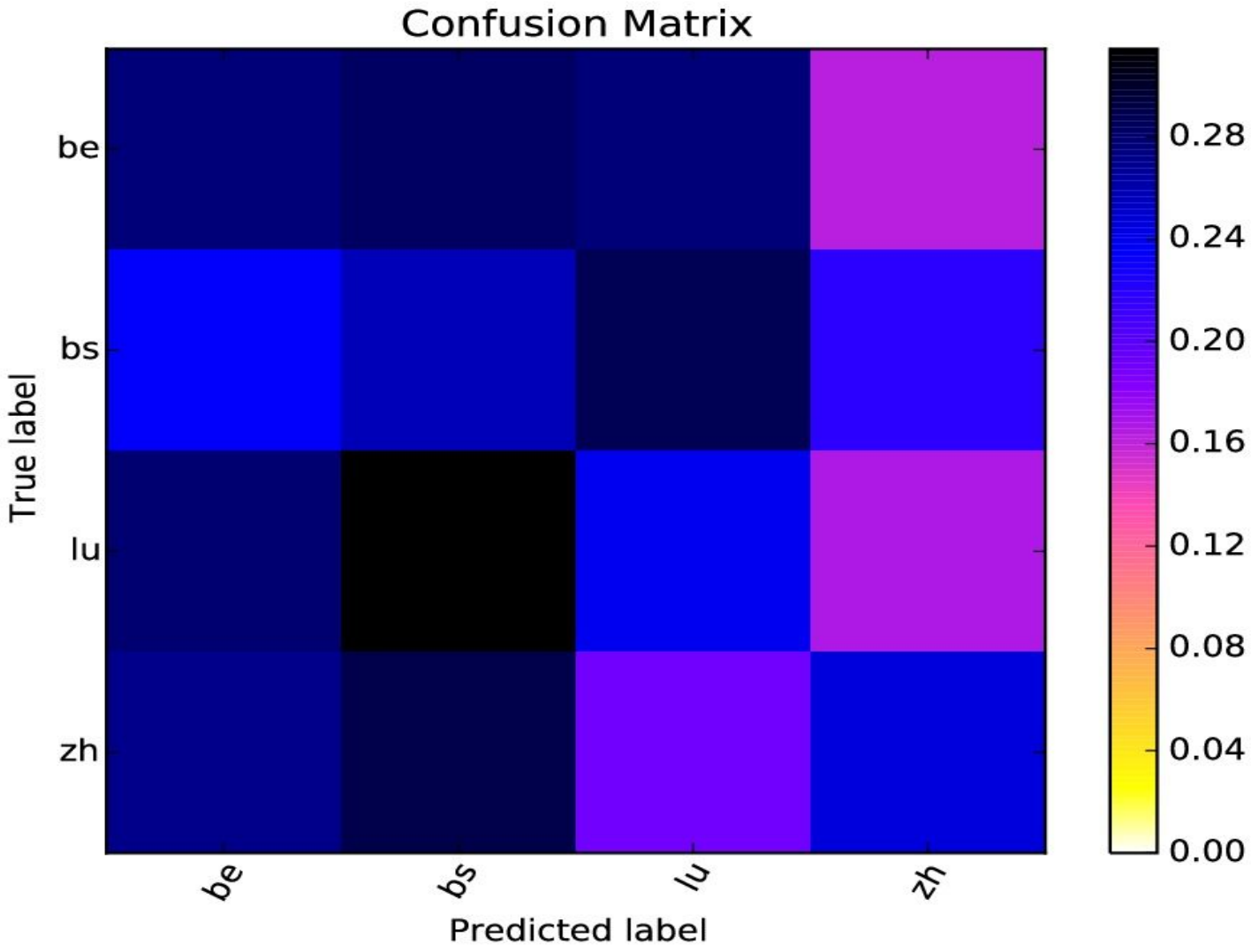

**Fig 8.** Confusion matrix for lexical features

## 7 Conclusion

This paper presents a deep learning based approach particularly LSTM and Bidirectional LSTM (B-LSTM) with lexical and bottleneck features of i-vector framework for recognizing Dialect Identification. 6 groups of similar languages are distinguished using LSTM architecture and the dialects of 5 regional languages in Arabic and 4 in German are identified by application of B-LSTM model. The i-Vector features give much better results in DI when compared to lexical

features.As a future work, the proposed model can be improved by hyperparameter tuning or by using hybrid neural networks.